\documentclass[letterpaper, 10 pt, conference]{ieeeconf}  

\IEEEoverridecommandlockouts                              

\usepackage{graphics} 
\usepackage[]{algorithm2e}
\usepackage{epsfig} 
\usepackage{mathptmx} 
\usepackage{amsmath} 
\usepackage{balance}

\title{\LARGE \bf
The Chef's Hat Simulation Environment for Reinforcement-Learning-Based Agents
}

\author{Pablo Barros, Anne C. Bloem, Inge M. Hootsmans, Lena M. Opheij, Romain H.A. Toebosch, \\Emilia Barakova and Alessandra Sciutti
\thanks{Pablo Barros, Anne C. Bloem, Inge M. Hootsmans, Lena M. Opheij, Romain H.A. Toebosch and Alessandra Siutti are with the Cognitive Architecture for Collaborative Technologies (CONTACT) Unit Istituto Italiano di Tecnologia, Genova, Italy , 
Emilia Barakova is with the Department of Industrial Design, University of Technology Eindhoven, Eindhoven, The Netherlands.
        {\tt\small pablo.alvesdebarros@iit.it}
}
}

\begin{document}

\maketitle
\thispagestyle{empty}
\pagestyle{empty}

\begin{abstract}

To achieve social interactions within Human-Robot Interaction (HRI)  environments is a very challenging task. Most of the current research focuses on Wizard-of-Oz approaches, which neglect the recent development of intelligent robots. On the other hand, real-world scenarios usually do not provide the necessary control and reproducibility which are needed for learning algorithms. In this paper, we propose a virtual simulation environment that implements the Chef's Hat card game, designed to be used in HRI scenarios, to provide a controllable and reproducible scenario for reinforcement-learning algorithms.

\end{abstract}

\section{Introduction}

Modeling the action-perception cycle within social robots, as a continuous adaptation mechanism, is one of the most important, and challenging, goals of human-robot interaction \cite{crossman2018influence}. Most of the social tasks robots are expected to perform in the near future demand not only the perception of human traits but also the understanding of such traits in a given context through the modulation of decision-making \cite{sandini2018social}.


In particular, the development of recent cognitive architectures to deal with social interactions became of great interest in recent years \cite{franklin2013lida,tanevska2018designing}. As affect is present in most human-human and also human-robot interactions, most of the social cognitive architectures involve or take into consideration affective content. To assess how others feel while we interact with them gives us important context on the situation we are in, but also on the appropriateness of our own actions \cite{hirokawa2018adaptive}. Taking this information as part of our decision-making, allow us to adapt properly to perceiving and understanding others, to improve the engagement and effectiveness of the interaction, and to guarantee that a certain message is passed through our actions. Such characteristics are extremely important when designing and developing robots for health-care, tutoring and even as specific companions.

 A common problem, however, arises when developing intelligent social robots based on human interaction. To design, evaluate and to validate such systems is a computationally expensive task. You need different persons, behaving in a different manner and in different situations, and usually, this scenario is not reproducible which increases the number of variables a researcher has to validate in order to derive grounded claims. Due to this limitation, most of the current solutions for social cognitive architectures in robots are based on interaction strategies focused on contagion \cite{van2018generic}, by repeating what was perceived or relied upon simple decision trees for generating behavior \cite{tuyen2018emotional}.

Such solutions are suited for the simple interaction scenarios in which they were deployed. The limitation on designing, implementing and deploying realistic interaction scenarios, in particular where affect plays a role, is one of the problems that must be solved before the deployment of such robots in real life. In a previous attempt without using a robot, a game specially designed to provoke to favor one and disfavor another player in a multiplayer interaction was shown to provoke emotions \cite{barakova2015automatic}, which lasted during the repetitive games between the same players \cite{gorbunov2017memory}. Adding a robot to such a scenario, however, would demand an incredible powerful simulation environment, or several rounds of real-life experiments in order to collect enough real-world data to adapt learning algorithms.

A common solution for such cases is the use of simulated environments \cite{brockman2016openai}. With the recent development of reinforcement learning algorithms, the construction and design of complex simulation environments flourished. Such environments allow fast-paced simulations of different tasks, the calculation of specific step-reward functions and are the basis for the recent groundbreaking applications of deep reinforcement learning. Most of these scenarios, however, are not developed with HRI in mind. Most of them simulate situations which are based on single agents \cite{shi2019pyrecgym}, classic reinforcement learning problems \cite{cullen2018active}, robotic simulations \cite{zamora2016extending}, or, more recently, playing video games \cite{torrado2018deep}.

To address the problem of providing a standard and easily reproducible affective interaction scenario for HRI, we recently developed and validated a novel card game \cite{barros2020food}. The game, named Chef's Hat, was designed with specific HRI requirements in mind, which allows it to be followed and modeled by artificial agents with ease. Also, the game mechanics were designed to evoke different affective interactions within the game, which can be easily perceived and displayed by a robot. Furthermore, the game elements were design to facilitate the extraction of the game state through the use of QR-codes and specific turn taking actions, which do not break the game flow.

To continue our efforts to facilitate the development of a reproducible affective action-perception cycle, we propose here a novel simulation environment, based on the OpenAI Gym toolkit \cite{brockman2016openai}, for the Chef's Hat card game. The simulation environment encapsulates all the complex rules and mechanics of the game and allows the training of multiple artificial agents as players. With it, we can simulate group dynamics with easy by the implementation of specific types of agents. 

The entire environment is portable and is freely available, to facilitate its dissemination. Besides training and evaluating artificial agents, the environment also can store different gameplays, to generate learning datasets used by the agents. It also can load specific gameplays and customize different agents with real-world experiences so as to have a controlled environment for focused experiments. It can also render played games, providing the necessary information for human-level analysis.

In this paper we formally describe the functioning of the environment, following all the game mechanics. We also present a set of baseline experiments to both demonstrate the environment but also to validate the capability of the environment to simulate the Chef's Hat card game. Our experiments involve the development and training of virtual agents, based on Deep Q-learning \cite{van2016deep}, and the analysis of how these agents behave while learning how to play the game. 



\section{Chef's Hat Simulation Environment}

Our HRI scenario is based on a card game played by four players. The card game scenario was chosen as our action-perception cycle development environment as it is a controllable situation, where each player has its turn to take specific actions, and yet provides the ground for natural interaction between the players. Within this scope, we needed a game that allowed the players to develop strategies while playing, and that, within the game mechanics, evoke the display of different affective states. Also, the evolution of the affective behavior of the players through the game should influence their strategy and decisions. 

\begin{figure}
\begin{center}
  \includegraphics[width=0.7\linewidth]{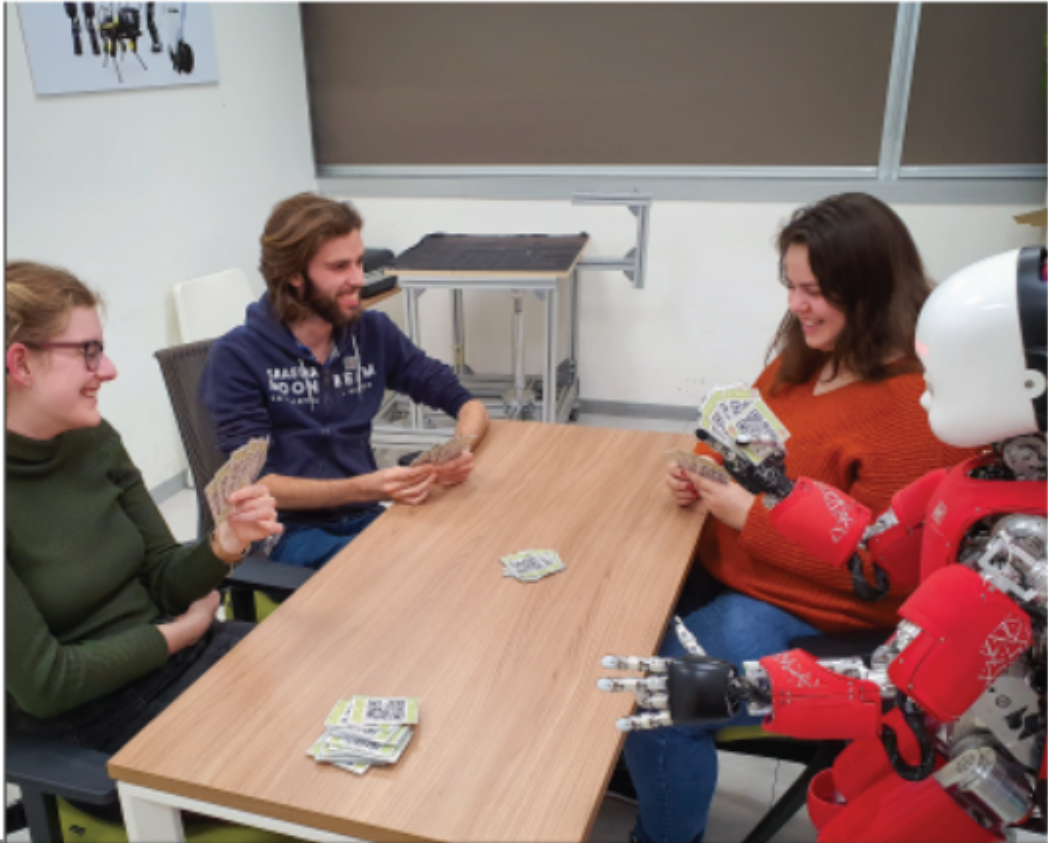}
\end{center}
  \caption{Illustration of the Chefs Hat card game been played by three humans and a robot.}
\label{fig:icubExample}
\end{figure}

In this regard, we recently developed and validate the Chef's Hat card game \cite{barros2020food}, illustrated in Figure \ref{fig:icubExample}. It includes complex game mechanics that were modeled in its entirety in our Gym environment \cite{brockman2016openai}. Below, we describe the game mechanics and the details of the environment functionalities. 

\subsection{Embedded Chef's Hat Mechanics}

The development of the Chef's Hat mechanics followed two main principles: 
1) To provide restricted, but natural interaction, where affective behavior naturally arises; 2) to provide turn-taking, i.e. an organized structure, in which a robot, such as the iCub \cite{metta2008icub}, has a supportive infrastructure and capacity to process incoming information and generate behavior without breaking the fluidity of the interaction.

The game simulates a kitchen, and it has a role-based hierarchy: each player can either be a Chef, a Sous-Chef, a Waiter, or a Dishwasher. The players try to be the first to get rid of their ingredient cards and become the Chef. This happens for multiple rounds (or Shifts, each of them detailed in Algorithm \ref{alg:ChefsHat}) until the first player reaches 15 points.

\begin{algorithm}
 Shuffle the deck; \\
Deal an equal amount of cards per player; \\
  Exchange roles; \\
  Exchange cards; \\
   \If{special action is evoked}
  {
    Do special action;
  }
  
 First player discard cards. \\
 
 \While{not end of the shift}{
 
  \For{ each player}{

      \eIf{player can, and want, to discard}{
       discard cards\;
       }{
        pass\;
      }
      \If{All players passed}
      {
        Clean the board;
      }
        \If{All players finished}
      {
        End of shift.
      }

   }
 }
 \caption{The playing flow of one Shift of the Chef's hat card game.}
 \label{alg:ChefsHat}
\end{algorithm}

As exhibited above, during every Shift there are three phases: Start of the Shift, Making Pizzas, End of the Shift.

At the start of the Shift, the cards are shuffled and dealt with the players. Then, the exchange of roles starts based on the previous Shift end positions. Who finished first becomes the Chef, who finishes second become the Sous-Chef, third the Waiter and fourth the Dishwasher. Once the roles are exchanged, the exchange of the cards starts. The Dishwasher has to give the two cards with the highest values to the Chef, who in return gives back two cards of their liking. The Waiter has to give their lowest valued card to the Sous-Chef, who in return gives one card of their liking. The change of roles is extremely necessary to change the game balance, rewarding the players who finished first in the last Shift and encouraging them to win the next one.

Once all the roles are finished, the players have the chance to do a special action. If a player has two jokers at the start of the Shift in its hand, they can choose to play their special action: in case of the Dishwasher this is "Food Fight" (the hierarchy is inverted), in case of the other roles it is "Dinner is served" (there will be no card exchange during that the Shift).

Then, the making of the pizzas starts. The person who possesses a Golden 11 card may start making the first pizza of the Shift. A pizza is prepared when ingredient cards are played on the pizza base on the playing field. A pizza is done when no one can (or wants to) put on any ingredients anymore. The rarest cards have the lowest numbers. A player can play cards by laying down their ingredient cards on the pizza base. To play cards, they need to be rarer (i.e. lowest face values) than the previously played cards. The ingredients are played from highest to the lowest number, so from 11 to 1. Players can play multiple copies of an ingredient at once, but always have to play an equal or greater amount of copies than the previous player did. If a player cannot (or does not want) to play, they pass until the next pizza starts. A joke card is also available and when played together with other cards, it assume their value. When played alone, the joker has the highest face value (12).

At the end of the Shift, the new roles are distributed among the players according to the order of finishing, and every player gets the number of points related to their role. The Chef gets 5 points, the Sous-Chef gets 3 points, the Waiter gets 1 point and the Dishwasher gets 0 points. The game continues until one of the players reaches 15 points.

\begin{figure}
    \centering
    \includegraphics[width=1.0\columnwidth]{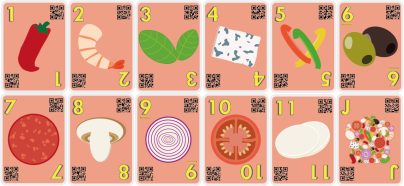}
    \caption{Ingredient cards, and the Joker, with their corresponding face number. The lower the number, the rarer the card.}
    \label{fig:cards}
\end{figure}

The ingredient cards, illustrated in Figure \ref{fig:cards}, needed to be easily recognizable, by the players and the robot, both when played on the playing field and when exchanged among players at the start of the Shift. For this, QR- codes were chosen, as they are hard for humans to recognize or memorize, but easy for the robot to read. Also, the QR-codes allow that a camera placed on top of the playing field capture the game-state and save it. This is extremely important when creating a learning database to be used together with the virtual environment.

\begin{figure}
    \centering
    \includegraphics[width=0.7\columnwidth]{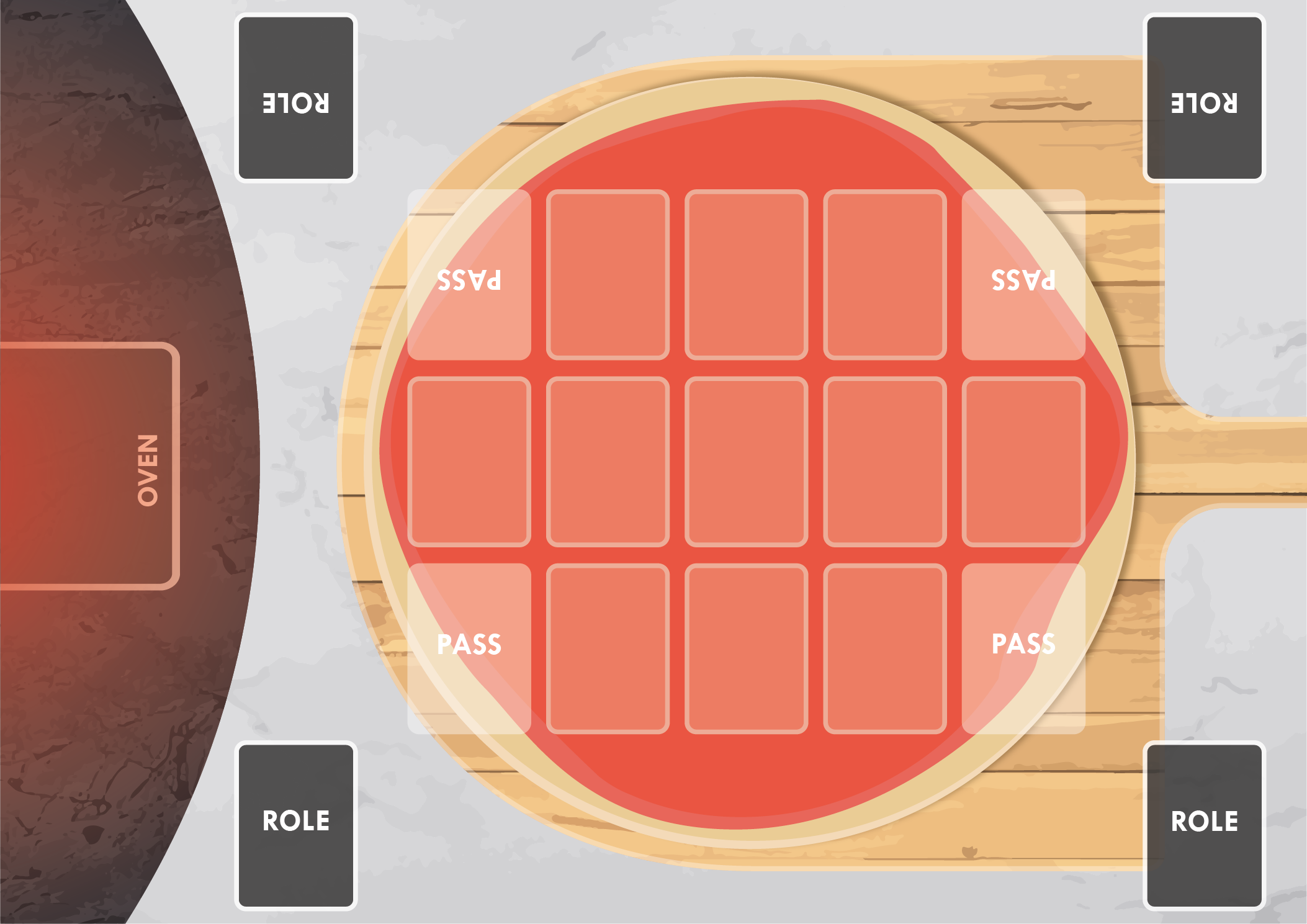}
    \caption{Playing field where the cards are placed, representing a pizza board.}
    \label{fig:field}
\end{figure}

The cards are to be placed on the playing field, illustrated in Figure \ref{fig:field}. To guarantee that the players lay down the cards without stacking them, and hindering them to be recognized by automatic systems,  we redesigned the playing field to have 11 different marked places in which players could place their cards on the pizza.

\subsection{OpenAI Gym Environment}

OpenAI Gym \cite{brockman2016openai} is a very popular toolkit that facilitates the development and the dissemination of simulation environments for training reinforcement learning agents. It enables the creation of standardized environments that allow the establishment of a set of specific rules for a simulation, the calculation of varied types of rewards and the logging and visualization of training artificial agents. Recently, several simulation environments were released using the OpenAI gym, which facilitates its reproducible and evaluation with different reinforcement learning algorithms.

We ported the Chef's Hat game into the OpenAI Gym and implemented all the complex game rules and mechanics. The environment is freely available\footnote{https://github.com/pablovin/ChefsHatGYM/blob/master/Readme.md}, and we hope that it helps us to standardize the learning of game strategies within our card game, but also to collect and share data for different learning algorithms.

Our environment is composed of a game handler, that deals with all the game mechanics. The game handler initiates the game, deal the cards, maintain all the stateful representations of the deck, the players' hands, the game field, and the score. It controls the Shifts and the turns, but most importantly the environment controls the actions each player performs.

The environment calculates the current game state by aggregating the current player's hand and the current cards on the board. Using this standard state representation we can give the learning agents the possibility to learn specific strategies purely based on the cards it holds and the cards which were displayed. Of course, as the environment is fully customizable, the current game state can be composed of any other variable which might help the agent to success in its task. 

Each action taken by an agent is validated based on a look-up-table, created on the fly based on the player's hand and the cards in the current playing field, to guarantee that a taken action is allowed given the game context. The look-up-table is extremely important as it guarantees that the game rules are maintained.

\begin{figure}
    \centering
    \includegraphics[width=0.78\columnwidth]{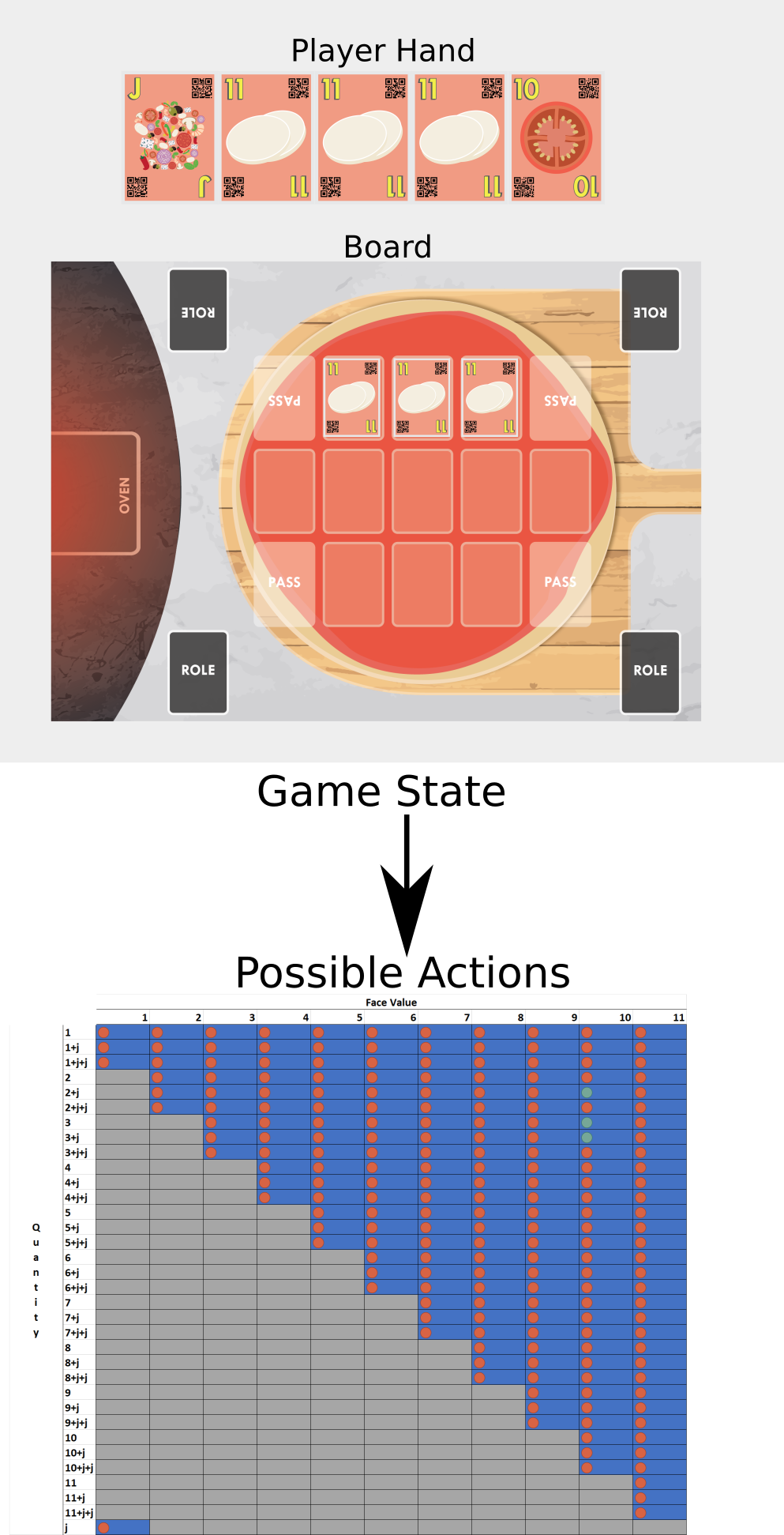}
    \caption{Example of possible actions given a certain game state. The columns represent the card face values, and the rows represent the number of cards to be discarded. The letter "j" represents the presence of a joker. The look-up table is created on the fly, and it marks all the actions which are allowed based on the game mechanics (blue regions) and the ones which are not allowed (the gray regions). For this given game state, the player would be allowed only to perform the actions marked with the green dots.}
    \label{fig:possibleActions}
\end{figure}

The actions are calculated based on the number of possibilities of the look-up table. The standard game, with a deck of 121 cards, has a total of 200 possible actions, which encapsulate all the possible moves a player can do: to discard one card of face value 1 represents one move or to discard 3 cards of face value 1 and a joker is another move, while passing is considered another move. Figure \ref{fig:possibleActions} illustrates an example of calculated possible actions given a game state. The blue areas mark all the possible action states, while the gray areas mark actions that are not allowed due to the game's mechanics. We observed that, given this particular game state, this player would only be allowed to perform one of three actions (marked in green), while any other action (marked in red) would be considered as invalid and not carried on.

The environment allows the customization of the game itself. We can easily choose how many players will be playing the game, how many cards a deck can have and how many of the playing agents are to be trained. The agents are also customized and follows a standard implementation protocol. This allows the implementation and deployment of a large variety of agents, from complex learning agents that might take external factors to learn the game strategy (e.g. from an external camera reading a real-game) to naive agents that do specific actions following simple rules.

For each action that an agent performs, the game environment calculates a specific reward. Again, as the environment is fully customizable, the reward calculation can be updated accordingly to the needs of the training agents. For example, giving the highest reward for an agent that performs a valid action - that means, an action which follows the game rules - can be used to train an agent to learn the rules of the game. Later on, this reward can be updated to make the agent learn how to win the game.

Another aspect of our environment is the logging of actions and states. It allows us to create snapshots of each played game, that can be used to create playing datasets which are extremely helpful for further training intelligent agents. Each step of the game play is recorded in a different set of files and can be retrieved later on with ease. 

The stored games can be used by the environment as modulation for specific agents to behave in a particular manner. That allows information obtained from real-world games, collected while real persons are playing it, to be easily inserted in the environment as primitives for the training the agents. The game status of the real games can be obtained via a single camera facing the playing field, and when saved in the same format as the one used by the environment, can be imported and used during the game.

\begin{figure}
    \centering
    \includegraphics[width=0.7\columnwidth]{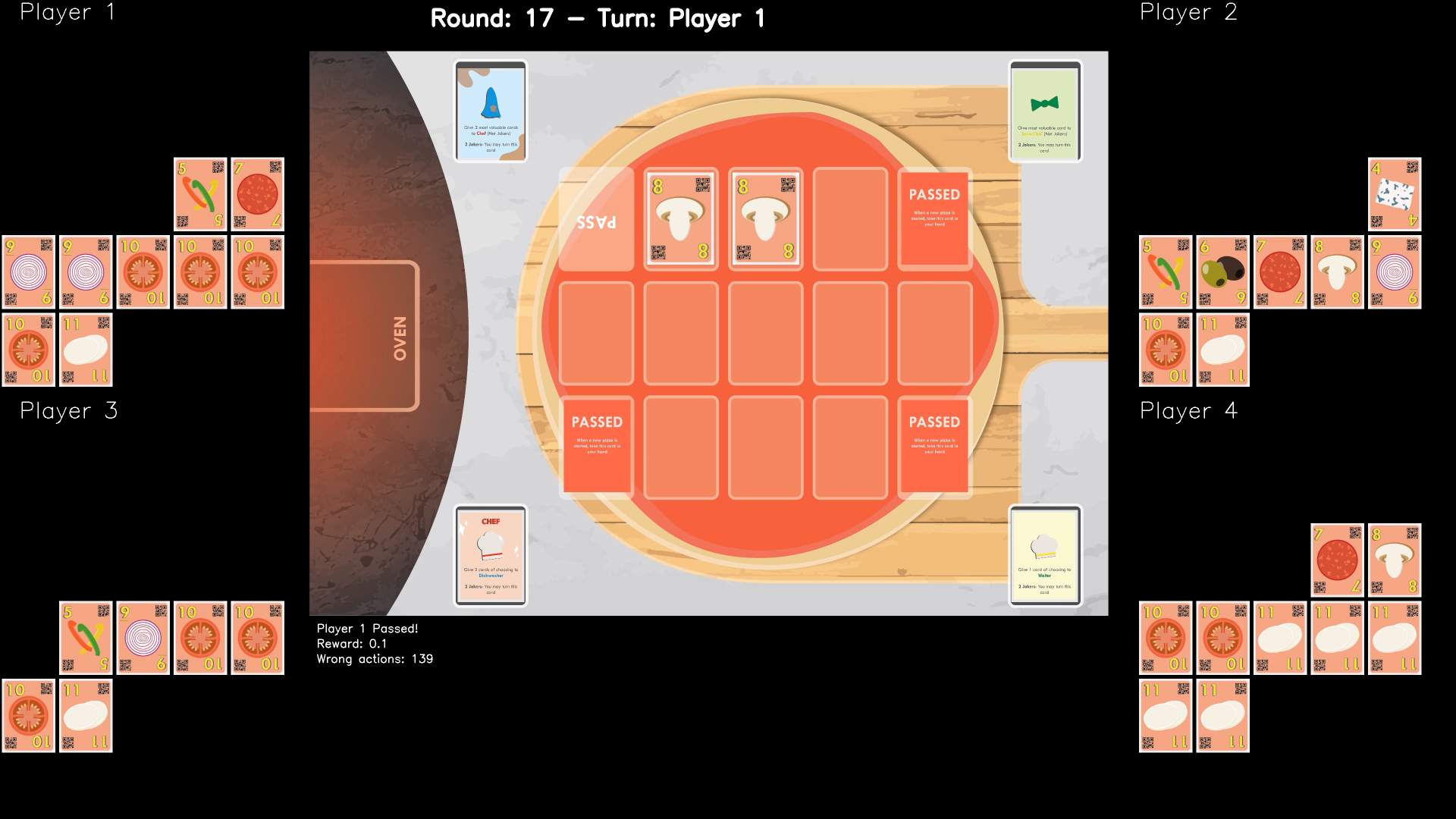}
    \caption{Example of the render function of the simulation environment.}
    \label{fig:simulationExample}
\end{figure}

Our environment also allows the rendering of played games. This function creates short videos of each played game, reproducing the actions of all players and the playing field. It is also an important function for a high-level analysis of the strategies taken by the learning agents, and also a simple and effective way to disseminate the learning of the agents. Figure \ref{fig:simulationExample} illustrates the rendering of a game \footnote{Full simulation available here: http://shorturl.at/ltAU0}.

\section{Validating the Chef's Hat Simulation Environment}

To validate how our environment implements the entire Chef's Hat game, and to demonstrate how useful it can be for training artificial agents, and as well to set the initial standardization for future research, we realized two types of experiments: first, we focus on teaching an agent the rules of the game, validating our implementation of the complex game mechanics of the Chef's Hat card game.
Second, we simulate the learning of an agent and teach it how to play the game. This experiment will serve as a demonstration of how to extract important information and agent's behavior from our environment.

\subsection{Implementing the Artificial Players}

We used two different types of agents in our experiments. The first agent runs completely random, without any learning at all. Every time it has to do an action, it chooses a random action which is them validated by the game environment. The random agent allows us to measure the baseline probabilities of the game itself and serves as a naive benchmark when playing against our learning agents.


The second agent presents a learning architecture based on the recent deep Q-learning algorithm \cite{schulman2017proximal}. Q-learning is an off-policy algorithm that learns to optimize a certain set of actions, which are outside the model's world, based on the quality of a taken action. The quality, Q, represents how useful the taken action is towards the end-goal of the algorithm. Deep Q-learning is an evolution of the standard Q-learning method and introduces two novel aspects: a target model and the experience replay. The target model helps to stabilize the learning of Q by providing a stable Q-estimation over the training. The experience replay stores the agent's own experience, by saving important steps taken by the agent, to increase the available data for learning state/action pairs. Deep Q-learning was recently applied to teach agents to play complex video games with great success \cite{hausknecht2015deep, hester2018deep, meng2019qualitative}. In our scenario, Q-learning has an advantage of updating the agent towards winning the game, and not only towards taking single-stepped actions.

The learning agent was optimized using a TPE optimization implemented by the Hyperopt \cite{bergstra2013hyperopt} library. Each of the learning agents implemented a single optimization routine for minimizing its learning loss. We implemented the agent using Keras library \cite{gulli2017deep}, and Table \ref{tab:parametersDQL} describes the agent final configuration and parameters.

\begin{table}[h!]
\centering
\begin{tabular}{ c | c | c }
 Layer & Neurons & Activation Function \\ \hline
 Dense layer & 128 & ReLU \\
 Dense layer & 256 & ReLU \\
 Dense layer & 200 & Tanh (output layer) \\\hline
 Training Parameter & \multicolumn{2}{c}{Value}  \\ \hline
 Batch Size & \multicolumn{2}{c}{ 32} \\
 Memory size & \multicolumn{2}{c}{ 2000} \\  
 discount rate & \multicolumn{2}{c}{ 0.95}  \\ 
 exploration decay & \multicolumn{2}{c}{ 0.995} \\\hline   
 
\end{tabular}
\caption{Parameters of the Deep Q-Learning agent used in our experiments.}
\label{tab:parametersDQL}
\end{table}


\subsection{Validation Procedures}

Our Experiment 1 focuses on obtaining the fundamental principles of the simulator. We run this experiment only with random players. We are interested on evaluating the impact that each of the game mechanics has on the chances of each random agent winning the game. We run specific games with and without the following mechanics: the presence of the joker, the exchange of cards in the beginning of each round and the presence of the special actions. 

Experiment 2 focuses on validating the simulation environment by teaching an agent how to learn the rules of the game. This experiment will demonstrate if the implemented rules allow the agent to reduce the number of invalid actions it takes while playing the game - and thus, learning the rules. This will be done by setting a reward system based on weather a taken action is valid or not. If the taken action is valid, the agent will receive maximal reward, whereas if the taken action is invalid, the agent will receive maximal punishment. We will use one Deep Q-Learning-based agent playing against three random agents, as this will allow us to properly evaluate if the agent is learning anything.


In Experiment 3, we will simulate a full-learning environment to demonstrate how we can collect and evaluate data from our simulation environment. We will use the trained agent from the first experiment an initial step, and we will train them to win the game. In this regard, the rewards will be updated to reflect the finishing position of each player and the number of cards each player has in its hand. By minimizing the number of cards in its hand, the agent learns that discarding cards is important, and combining this with the finishing position will direct the agent to win the game.  Table \ref{tab:rewardCalculation} illustrates the rewards chosen for each action. In this experiment we run the two extreme conditions: one learning agent against three random agents, and four learning agents against each other. This way we can evaluate the effect that an agent has on the learning of another one.

\begin{table}[h!]
\centering
\begin{tabular}{ c | c }
 Action & Reward \\ \hline
 \multicolumn{2}{c}{Experiment 2} \\\hline
 Valid action & 1.0 \\  
 Invalid action & -1  \\  
  \multicolumn{2}{c}{Experiment 3} \\\hline
 Valid Action &  
    $\left (\frac{\left ( 1-cardsHand \right ) \times 100}{initialCardsHand}\times 0.01 \right )*0.7$
   \\
  Finish Game & 
  $\left ( 1-finishingPosition \times 0.3 \right ) \times 0.3 $\\
 Invalid action & -1  \\ 
\end{tabular}
\caption{Reward calculation for specific actions taken by the agents in our experiments.}
\label{tab:rewardCalculation}
\end{table}

\subsection{Metrics}

For each of our experiments, we run 250 games. We then calculate specific metrics for each experiment.

In Experiment 1 we calculate the chances of the player starting the Shift to win that shift, in order to evaluate the impact of each of the observed mechanics. We also calculate the average number of rounds each game takes, to measure how these mechanics affect the games' length. 

For Experiments 2 and 3, in each game, we train the agents and we measure the average reward, the number of wrong actions (actions which are not allowed based on the game rules) per round, and the number of victories for each of the agents. 
The average reward will indicate us if each agent is learning how to optimize the actions through the game, while the number of wrong actions will indicate if the agent learns how to follow the rules of the game. The number of victories, finally, will indicate us if the agent is learning how to win the games. By measuring these three variables we can clearly see the effectiveness of each of the agents towards specific goals.

Also to evaluate the importance of the game mechanics towards winning the game, we calculated the relationship between starting a Shift and winning that same shift. This indicates how the game mechanics direct the learning of an agent and help us to validate if the chosen rewards allow the agent to win the game.

\section{Validation Results}

\subsection{Experiment 1: How each specific game mechanics impact the gameplay?}

Running the 250 games allowed us to calculate the impact of each of the observed mechanics. Table \ref{tab:Experiment1Results} report the final results. We observed that the initial condition, with all the mechanics, has the chance of a player winning the round if it starts the shift at 28\%. That means the original game mechanics favor slightly the player starting the round. When removing the joker, and with a stronger effect, the exchange of cards at the beginning of each Shift, we see an increase in the chances of the player starting the round. These results corroborate with what we observed from the real-game \cite{barros2020food} and demonstrate the importance of these mechanics. The special actions, although not contributing to the game's balance, introduced the affective factor within the players' interaction which is the basis of Chef's Hat game design.

The absence of the Joker also affected the number of rounds, but this is expected because without the Joker we have less cards in the deck. Removing the exchange cards, however, reduced the amount of rounds, which is mostly connected to the increase of chances of the player starting the Shift to win it. If the starting player always win, the number of rounds necessary to this happen dropped.

\begin{table}[h!]
\centering
\begin{tabular}{ c | c | c }
 Mechanic & Start/Winning Shift & Rounds \\ \hline
 
 All & 28\% & 35 \\
 Without Joker & 35\%  & 33\\
 Without Exchange cards & 60 \% & 31\\ 
 Without Special actions & 28\% & 35\\
 
\end{tabular}
\caption{Results of the average chances of a player starting the shift wins the game and number of rounds when removing each of the specific mechanics.}
\label{tab:Experiment1Results}
\end{table}

\subsection{Experiment 2: Can an agent learn the rules of the game?}

\begin{figure*}[h!]
    \centering
    \includegraphics[width=2.0\columnwidth]{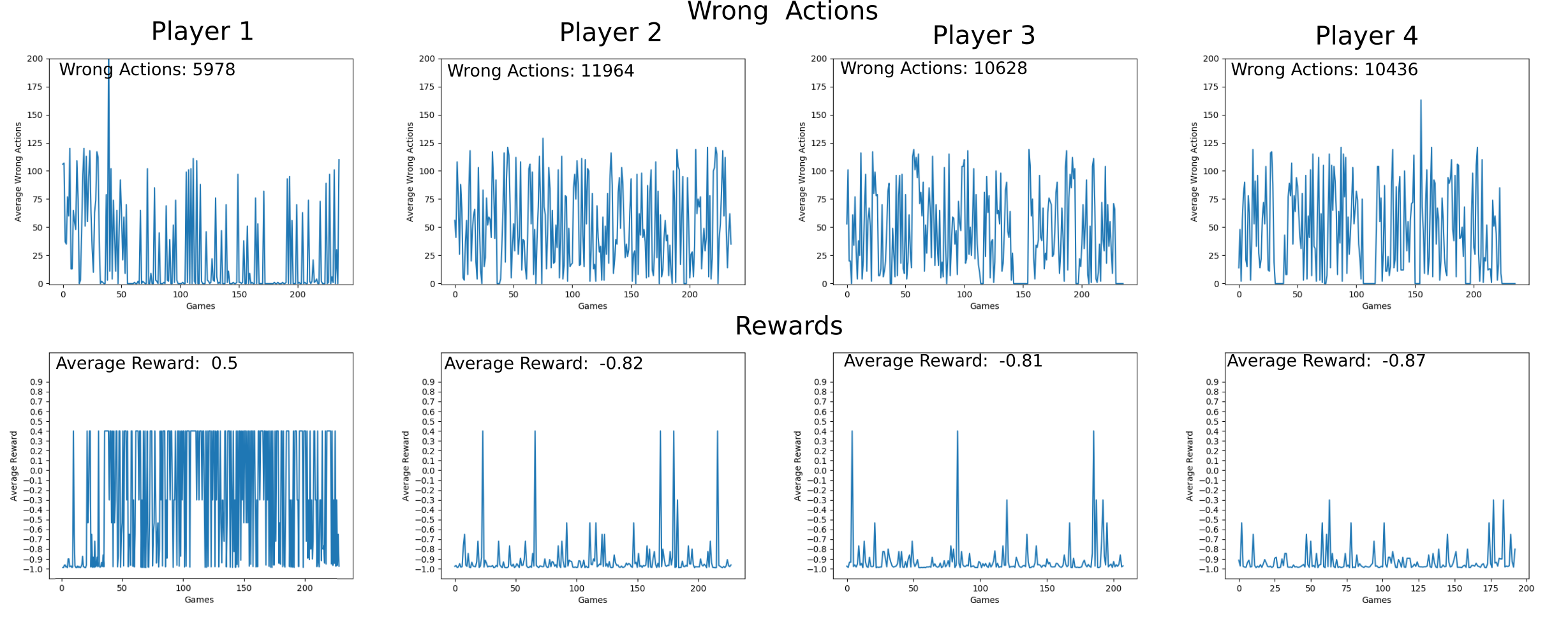}
    \caption{Wrong actions and average rewards per player for our first experiment, focusing on learning the rules of the game. A learning agent, Player 1, playing the game against three random agents.}
    \label{fig:firstExperiment}
\end{figure*}

After training one agent for 250 games against three other random agents, we observed an increase on the average reward and a decrease in the absolute number of wrong actions. Figure \ref{fig:firstExperiment} illustrates the result of training Player 1 with the Deep Q-Learning agent while playing against three other random agents. The absolute wrong actions was almost half of the random agents, while the average reward was 0.5 while the best random agent had an average reward of -0.82.  Also, observing the average reward of Player 1, we can see that it has a very high variation when compared to the random agents. This is probably due to Deep Q-Learning exploration mechanism. We observed the reward steadily changing between minimum and maximum over time.

While our simulation shows that the agent trained with Deep Q-Learning was able to learn the game rules, it did not improve the amount of victories for Player 1. In our 100 simulations, Player one won 65 of them, while Player 2 won 62, Player 3 won 63 and Player 4 won 60 simulations.

\subsection{Experiment 3: Can an agent learn how to win the game?}

We run our third experiment based on two conditions: first, we take the trained agent, Player 1, from Experiment 2 and put it to play against three other random agents. Figure \ref{fig:secondExperiment_1} illustrates our results. We observe that the average reward increased drastically and fast for Player 1, which indicates that it learned how to play the game in a fast-pace. Our final victories count show that Player 1 won a total of 215 games, in fact after the 50th game, player one won all the simulations. The reward pattern of Player one over the games display that it took some rounds to learn a winning strategy. Once it learned, around the 50th game, Player one achieve a much higher and steady reward than the other players.

\begin{figure*}
    \centering
    \includegraphics[width=2.0\columnwidth]{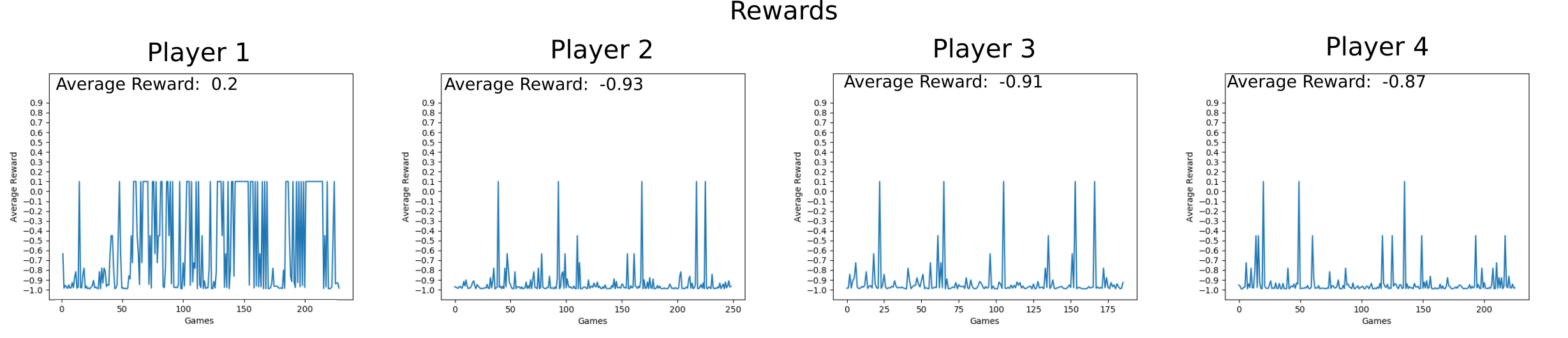}
    \caption{Average rewards per player for our second experiment, focusing on winning the game. A learning agent, Player 1, playing the game against three random agents.}
    \label{fig:secondExperiment_1}
\end{figure*}

Our second condition had four trained agents playing against each other. This experiment illustrates how the Deep Q-Learning algorithm can deal with an extremely changing environment. In this experiment we observed that the reward differences between the players was much smaller. Player 1 ended up winning 117 games, while Player 2 won 45, Player 3 won 55 and Player 4 won 33 of the simulations.

\begin{figure*}
    \centering
    \includegraphics[width=2.0\columnwidth]{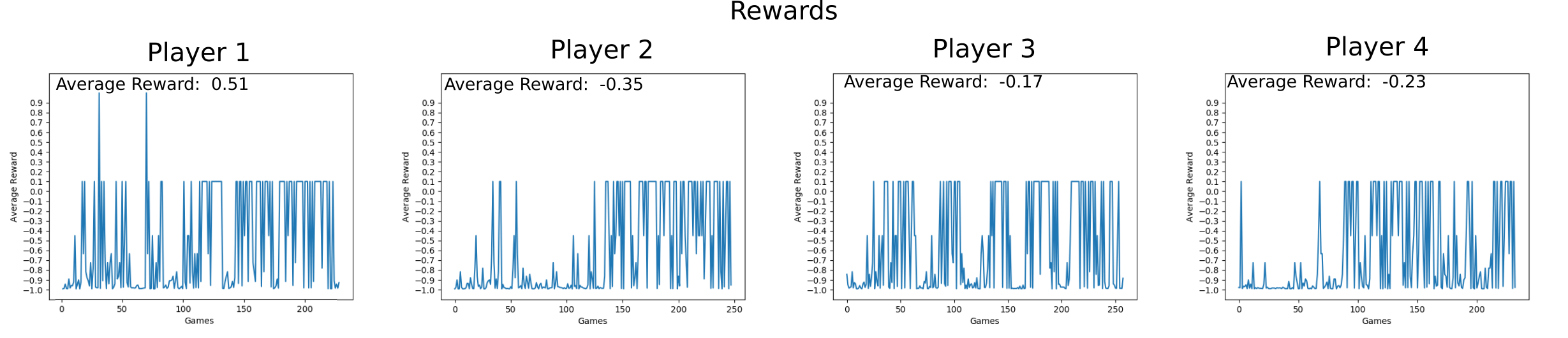}
    \caption{Average rewards per player for our second experiment, focusing on winning the game. Four learning agents playing the game against each other.}
    \label{fig:secondExperiment_1}
\end{figure*}

Observing the reward plot we can see that Player 1 started with much higher rewards than the others, while Player 2 took more than 100 games to start receiving a higher reward. We observe that Player 1 won more games probably due to his ability to create an early strategy to win, while the others had to adapt their game-style to beat Player one strategy. 

We also calculate the correspondence between starting a Shift and winning that same Shift. For the fully random agents, this value was around 30\%.


\section{Discussions}


Our experiments serve first as a validation for our Chef's Hat simulation environment, but also to illustrates the capability of our environment to provide a rich set up for such agents to learn a competitive task. We envision that these results serve as inspiration for developing and applying other learning strategies to our environment. They also helped to respond to our main research question, on how to provide an environment that allows the simulation of a complex group-based competitive action following a real-world task with the same mechanics and rules.

Our first experiment demonstrates that the game mechanics do affect the behavior of the agents. We calculated how the specific mechanics affect the chances of an agent to win the game, which helps to validate that our environment do model the real-world dynamics we found when playing the game with real-persons. Also, our second and third experiment help us to validate how different learning agents can learn the game mechanics, and can learn how to win a game when playing with agents performing random moves. This is important to both confirm that indeed the environment can be used to train such agents, but also to provide initial research on how learning agents behave in our environment.

Although our second experiment showed that the learning agent was not able to produce a full-autonomous behavior without making any invalid action, the third experiment demonstrates that the agent still can learn how to win the game. We expect that a deeper study on the development of the reward function used to train the agents can help us to mitigate this behavior, as an agent that is trained to simulate a real-world robot playing the game cannot make mistake while playing, otherwise it can disrupt the game while playing with other humans.

So far we focused on the game strategy, with the initial aim of enabling an agent to autonomously learn to play and win, purely basing its decisions on the game status. However, the social card game considered was purportedly designed to generate also affective feedback from the players, which is expected to play a relevant role in decision making. The simulation environment has been designed with the possibility to deal with the affective dimension too. For example, combining the game status with affective representation or time taken to do an action, can allow the agent to take into consideration these characteristics when deciding for an action.

As the entire gameplay can be recorded, and easily re-played or incorporated into the agents' behavior, our environment allows for teaching physical robots to play the game through the run of endless virtual simulations. By having the robot to act as an agent in the simulator will allow it to integrate its real-world sensing capabilities with the environment's simulations. It can use it to create mental states of a current game or to evaluate novel strategies based on what the robot is currently perceiving. This way, the robot will be able to interact in the social game with human partners, enacting the decisions derived from the simulations and updating them as a function of the responses recorded by the human opponents. This will allow the robot to autonomously select the most appropriate action during the game, potentially both in terms of strategy and movement expressivity.




\section{Conclusion}

In this paper, we presented the first version of the novel Chef's Hat simulation environment. Our environment implements, using the OpenAI Gym toolkit, all the rules and game mechanics of the Chef's Hat HRI game. The game was designed to allow affective-aware interactions between humans and robots, and our simulation environment is a fundamental step towards allowing the development, evaluation, and dissemination of artificial agents to support social robots.

Our next steps include the inclusion of different learning agents into the environment, as well as the necessary tools to use the environment with human agents. We also envision the adaptation of the environment to receive live-feed data from a robot that is playing the game, such as audio and visual information, to run simulations while it plays the game with other humans.

Already in the current status, our contribution guarantees that the game simulation environment, being openly accessible, could serve as a shared tool for the community to investigate and model affect- and strategy- based human-robot interaction in groups.


\balance
\bibliographystyle{IEEEtran}
\bibliography{bib}

\end{document}